\documentclass{article} 
\usepackage{nips13submit_e,times}
\usepackage{url}
\usepackage{epsfig}
\usepackage{graphicx}
\usepackage{amsmath}
\usepackage{amssymb}
\usepackage{multirow} 
\usepackage{color}

\definecolor{darkblue}{RGB}{47,1,154}
\definecolor{darkred}{RGB}{240,40,40}
\definecolor{darkgreen}{rgb}{0,.6,0}

\title{Unsupervised feature learning by augmenting single images}

\author{
Alexey Dosovitskiy, Jost Tobias Springenberg and Thomas Brox\\
Department of Computer Science\\
University of Freiburg\\
79110, Freiburg im Breisgau, Germany \\
\texttt{\{dosovits,springj,brox\}@cs.uni-freiburg.de} \\
}

\graphicspath{{./images/}}

\newcommand{\cT}{\mathcal{T}}

\nipsfinalcopy 

\begin{document}

\maketitle

\begin{abstract}
When deep learning is applied to visual object recognition, data augmentation is often used to generate additional training data without extra labeling cost. It helps to reduce overfitting and increase the performance of the algorithm. In this paper we investigate if it is possible to use data augmentation as the main component of an unsupervised feature learning architecture. To that end we sample a set of random image patches and declare each of them to be a separate single-image surrogate class. We then extend these trivial one-element classes by applying a variety of transformations to the initial 'seed' patches. Finally we train a convolutional neural network to discriminate between these surrogate classes. The feature representation learned by the network can then be used in various vision tasks.  We find that this simple feature learning algorithm is surprisingly successful, achieving competitive classification results on several popular vision datasets (STL-10, CIFAR-10, Caltech-101). 

\end{abstract}

\section{Introduction}
Deep convolutional neural networks trained via  backpropagation have recently been shown to perform well on image classification tasks containing millions of images and thousands of categories \cite{Krizhevsky_NIPS2012, Zeiler_arxiv2013}. While deep convolutional neural networks have been known to yield good results on supervised image classification tasks such as MNIST for a long time  \cite{LeCun_IEEE1998}, the recent successes are made possible through optimized implementations, efficient model averaging and data augmentation techniques~\cite{Krizhevsky_NIPS2012}. The feature representation learned by these networks achieves state of the art performance not only on the classification task the network is trained for, but also on various other computer vision tasks, for example:  classification on Caltech-101~\cite{Zeiler_arxiv2013, Donahue_arxiv2013}, Caltech-256~\cite{Zeiler_arxiv2013}, Caltech-UCSD birds dataset~\cite{Donahue_arxiv2013}, SUN-397 scene recognition database~\cite{Donahue_arxiv2013}; detection on PASCAL VOC dataset~\cite{Girschik_arxiv13}. This capability to generalize to new datasets indicates that supervised discriminative learning is currently the best known algorithm for visual feature learning. The downside of this approach is the need for expensive labeling, as the amount of required labels grows quickly the larger the model gets. For this reason unsupervised learning, although currently underperforming, remains an appealing paradigm, since it can make use of raw unlabeled images and videos which are readily available in virtually infinite amounts. 

In this work we aim to combine the power of discriminative supervised learning with the simplicity of unsupervised data acquisition. The main novelty of our approach is the way we obtain training data for a convolutional network in an unsupervised manner. In the standard supervised setting there exists a large set of labeled images, which may be further augmented by small translations, rotations or color variations to generate even more (and more diverse) training data. 

In contrast, our method does not require any labeled data at all: we use the augmentation step alone to create surrogate training data from a set of unlabeled images. We start with trivial surrogate classes consisting of one random image patch each, and then augment the data by applying a random set of transformations to each patch. After that we train a convolutional neural network to classify these surrogate classes. The feature representation learned by the network is, by construction, discriminative and at the same time invariant to typical data transformations. Nevertheless it is not immediately clear: Would the feature representation learned from this surrogate task perform well on general image classification problems? Our experiments show that, indeed, this simple unsupervised feature learning algorithm achieves competitive or state of the art results on several benchmarks.


By performing image augmentation we provide prior knowledge about natural image distribution to the training algorithm. More precisely, by assigning the same label to all transformed versions of an image patch we force the learned feature representation to be invariant to the transformations applied. This can be seen as an indirect form of supervision: our algorithm needs some expert knowledge about which transformations the features should be invariant to. However, similar expert knowledge is used in most other unsupervised feature learning algorithms. Features are usually learned from small image patches, which assumes translational invariance. Turning images to grayscale assumes invariance to color changes. Whitening or contrast normalization assumes invariance to contrast changes and, largely, color variations.


\subsection{Related work}
Our approach is related to a large body of work on unsupervised
learning and convolutional neural networks. In contrast to our
method, most unsupervised learning
approaches, e.g.~\cite{Hinton_NC2006, Hinton_Science2006,
  Vincent_ICML2008, Coates_NIPS2011, Zou_NIPS2012}, rely on modeling
the input distribution explicitly~-- often via a reconstruction error term~--
rather than training a discriminative model and thus cannot be used to
jointly train multiple layers of a deep neural network in a
straightforward manner. Among these unsupervised methods, most
similar to our approach are several studies on learning
invariant representations from transformed input samples, for
example~\cite{Sohn_ICML2012, Zou_NIPS2012, Hui_ICML2013}. 

Our proposed method can be related to work on metric learning, for example~\cite{Goldberger_NIPS2004, Hadsell_CVPR2006}. However, instead of enforcing a metric on the feature representation directly, as in~\cite{Hadsell_CVPR2006}, we only implicitly force the representation of transformed images to be mapped close together through the introduced surrogate labels. This enables us to use discriminative training for learning a feature representation which performs well in classification tasks. 

Learning invariant features with a discriminative objective was
previously considered in early work on tangent propagation~\cite{Simard1992}, which aims to learn  features invariant to small predefined transformations by directly penalizing the derivative of the network output with respect to the parameters of the transformation. In contrast to their work, our algorithm does not rely on labeled data and is less dependent on a small magnitude of the applied transformations. Tangent propagation has been successfully combined with an unsupervised feature learning algorithm in \cite{Rifai-nips-2011} to build a classifier exploiting information about the manifold structure of the learned representation. This, however, again comes with the disadvantages of reconstruction-based training.

Loosely related to our work is research on using unlabeled data for regularizing supervised algorithms, for example self-training~\cite{Amini_ECAI2002} or entropy regularization~\cite{Grandvalet_SSL2006,Lee_ICML_WCLR2013}. In contrast to these semi-supervised methods, our training procedure, as mentioned before, does not make any use of labeled data. Finally, the idea of creating a pseudo-task to improve the performance of a supervised algorithm is used in~\cite{Ahmed_ECCV2008}. 

\section{Learning algorithm}
Here we describe in detail our feature learning pipeline. The two main stages of our approach are generating the surrogate training data and training a convolutional neural network using this data.

\subsection{Data acquisition}
\begin{figure}[t]
\centering
\begin{minipage}[t]{.5\textwidth}
  \centering
  \includegraphics[width=.95\linewidth]{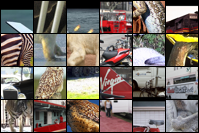}
  \begin{minipage}[t]{.9\linewidth}
  \caption{\label{fig:sample_patches}Random patches sampled from the STL-10 unlabeled dataset which are later augmented by various transformation to obtain surrogate classes for the neural network training.}
  \end{minipage}
\end{minipage}%
\begin{minipage}[t]{.5\textwidth}
  \centering
  \includegraphics[width=.95\linewidth]{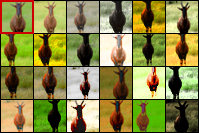}
  \begin{minipage}[t]{.9\linewidth}
  \caption{\label{fig:sample_transformations}Random transformations applied to one of the patches extracted from the STL-10 unlabeled dataset. Original patch is in the top left corner.}
  \end{minipage}
\end{minipage}
\end{figure}

The input to our algorithm is a set of unlabeled images, which come
from roughly the same distribution as the images we later aim to
classify. We randomly sample $N \in [50,\, 32000]$ random patches of
size $32 \times 32$ pixels from different images, at varying positions
and scales. We only sample from regions with considerable gradient
energy to avoid getting uniformly colored patches. Then we apply $K
\in [1,\,100]$ random transformations to each of the sampled
patches. Each of these random transformations is a composition of four
random 'elementary' transformations from the following list:
\begin{itemize}
 \item Translation: translate the patch by a distance within $0.25$ of the patch size vertically and horizontally.
 \item Scale: multiply the scale of the patch by a factor between $0.7$ and $1.4$. 
 \item Color: multiply the projection of each patch pixel onto the principal components of the set of all pixels by a factor between $0.5$ and $2$ (factors are independent for each principal component and the same for all pixels within a patch).
 \item Contrast: raise saturation and value (S and V components of the HSV color representation) of all pixels to a power between $0.25$ and $4$ (same for all pixels within a patch).
\end{itemize} 
We do not apply any preprocessing to the obtained patches other than subtracting the mean of each pixel over the whole training dataset. Examples of patches sampled from the STL-10 unlabeled dataset are shown in Fig.~\ref{fig:sample_patches}. Examples of transformed versions of one patch are shown in Fig.~\ref{fig:sample_transformations}.

\subsection{Training}
As a result of the procedure described above, to each patch $x_i \in X$ from the set of initially sampled patches $X = \{x_1, \ldots x_N \}$ we apply a set of transformations $\cT_i = \{ T_i^1, \ldots, T_i^K \}$ and get a set of its transformed versions $S_{x_i} = \cT_i x_i = \{T_i^j x_i|\, T_i^j \in \cT_i\}$. We then declare each of these sets to be a class by assigning label $i$ to the class $S_{x_i}$ and train a convolutional neural network to discriminate between these surrogate classes. Formally, we minimize the following loss function:
\begin{equation}
 L(X) = \sum\limits_{x_i \in X} \sum\limits_{T_i^j \in \cT_i} l(i,\, T_i^j x_i),
\end{equation}
where $l(i,\, T_i^j x_i)$ is the loss on the sample $T_i^j x_i$ with (surrogate) true label $i$. We use a convolutional neural network with cross entropy loss on top of the softmax output layer of the network, hence in our case
\begin{equation}
 l(i,\, T_i^j x_i) = CE(e_i, f(T_i^j x_i)), \quad CE(y,f) = - \sum\limits_{k} y_k \log f_k ,  
\end{equation}
where $f$ denotes the function computing the values of the output layer of the neural network given the input data, and $e_i$ is the $i$th standard basis vector.

For training the network we use an implementation based on the fast convolutional neural network code from~\cite{Krizhevsky_NIPS2012}, modified to support dropout. We use a fixed network architecture in all experiments: $2$ convolutional layers with $64$ filters of size $5 \times 5$ each followed by $1$ fully connected layer of $128$ neurons with dropout and a softmax layer on top. We perform $2\times2$ max-pooling after convolutional layers and do not perform any contrast normalization between layers. We start with a learning rate of $0.01$ and gradually decrease the learning rate during training. That is, we train until there is no improvement in validation error, then decrease the learning rate by a factor of $3$, and repeat this procedure several times until there is no more significant improvement in validation error.

\subsubsection{Pre-training}
In some of our experiments, in which the number of surrogate classes is large relative to the number of training samples per surrogate class, we observed that during the training process the training error does not significantly decrease compared to initial chance level. To alleviate this problem, before training the network on the whole surrogate dataset we pre-train it on a subset with fewer surrogate classes, typically $100$. We stop the pre-training as soon as the training error starts falling, indicating that the optimization found a direction towards a good local minimum. We then use the weights learned by this pre-training phase as an initialization for training on the whole surrogate dataset.

\subsection{Testing}
When the training procedure is finished, we apply the learned feature representation to classification tasks on 'real' datasets, consisting of images which may differ in size from the surrogate training images. To extract features from these new images, we convolutionally compute the responses of all the network layers except the top softmax and form a 3-layer spatial pyramid of them. We then train a linear support vector machine (SVM) on these features. We select the hyperparameters of the SVM via crossvalidation.

\section{Experiments}
We report our classification results on the STL-10, CIFAR-10 and Caltech-101 datasets, approaching or exceeding state of the art for unsupervised algorithms on each of them. We also evaluate the effects of the number of surrogate classes and the number of training samples per surrogate class in the training data. For training the network in all our experiments we generate a surrogate dataset using patches extracted from the STL-10 unlabeled dataset.

For STL-10 we use the usual testing protocol of averaging the results over 10 pre-defined folds of training data and report the mean and the standard deviation. For CIFAR-10 we report two results: 'CIFAR-10' means training on the whole CIFAR-10 training set and 'CIFAR-10-reduced' means the average over 10 random selections of 400 training samples per class. For Caltech-101 we follow the usual protocol with selecting 30 random samples per class for training and not more than 50 training samples per class for testing, repeated 10 times.

\subsection{Classification results}
In Table~\ref{tbl:classification} we compare our classification
results to other recent work. Our network is trained on a surrogate
dataset with $8000$ surrogate classes containing $150$ samples
each. We remind that for extracting features during test time we use
the first $3$ layers of the network with $64$, $64$ and $128$ filters respectively. The feature representation is hence considerably more compact than in most competing approaches. We do not list the results of supervised methods on CIFAR-10 (the best of which currently exceed $90\%$ accuracy), since those are not directly comparable to our unsupervised feature learning method. 

As can be seen in the table, our results are comparable to state of the art on CIFAR-10 and exceed the performance of many unsupervised algorithms on Caltech-101. On STL-10 for which the image distribution of the test dataset is closest to the surrogate samples our algorithm reaches $67.4\% \pm 0.6 \%$ accuracy  outperforming all other approaches by a large margin.

\begin{table}
  \begin{minipage}{\textwidth}
  \setcounter{mpfootnote}{\value{footnote}}
  \renewcommand{\thempfootnote}{\arabic{mpfootnote}}
    \small{
      \hspace*{-15pt}\begin{tabular}{|l|c|c|c|c|}
      \hline
                                                              &  STL-10                 &     CIFAR-10-reduced    &     CIFAR-10     &      Caltech-101         \\ \hline
      K-means~\cite{Coates_NIPS2011}                          &  $60.1 \pm 1$           &  $70.7 \pm 0.7$         &    $82.0$        &        ---              \\ \hline
      Multi-way local pooling~\cite{Boureau_ICCV2011}         &     ---                 &       ---               &      ---         &  $77.3 \pm 0.6$         \\ \hline
      Slowness on videos~\cite{Zou_NIPS2012}                  &  $61.0$                 &       ---               &      ---         &      $74.6$             \\ \hline
      Receptive field learning~\cite{Jia_CVPR2012}            &     ---                 &       ---               & $[83.11]$\footnotemark[1] & $75.3 \pm 0.7$ \\ \hline
      Hierarchical Matching Pursuit (HMP)~\cite{Bo_ISER2012}  &  $64.5 \pm 1$           &       ---               &      ---         &        ---              \\ \hline
      Multipath HMP~\cite{Bo_CVPR2013}                        &     ---                 &       ---               &      ---         & $\mathbf{82.5 \pm 0.5}$ \\ \hline
      Sum-Product Networks~\cite{Gens_NIPS2012}               &  $62.3 \pm 1$           &       ---               &    $[83.96]$\footnote[1]{As mentioned, we do not compare to the methods which use supervised information for learning features on the full CIFAR-10 dataset}     &        ---              \\ \hline
      View-Invariant K-means~\cite{Hui_ICML2013}              &  $63.7$                 & $\mathbf{72.6 \pm 0.7}$ & $\mathbf{81.9}$  &        ---              \\ \hline
      This paper                                              & $\mathbf{67.4 \pm 0.6}$ &  $69.3 \pm 0.4$         &    $77.5$        & $76.6 \pm 0.7$~\footnote[2]{There are two ways to compute the accuracy on Caltech-101: simply averaging the accuracy over the whole test set or calculating the accuracy for each class separately and then averaging these values. These methods differ because for many classes less than $50$ test samples are available. It seems that most researchers in the machine learning field use the first method, which is what we report in the table. When using the second method, our performance drops to $74.1 \% \pm 0.6 \%$}    \\ \hline
      
      \end{tabular}
      \caption{Classification accuracy on several popular datasets (in $\%$).}
      \label{tbl:classification}
    }
  \end{minipage}
\end{table}

\subsection{Influence of the data acquisition on classification performance}
\begin{figure}
\centering
\begin{minipage}[t]{.5\textwidth}
  \centering
  \includegraphics[width=.95\linewidth]{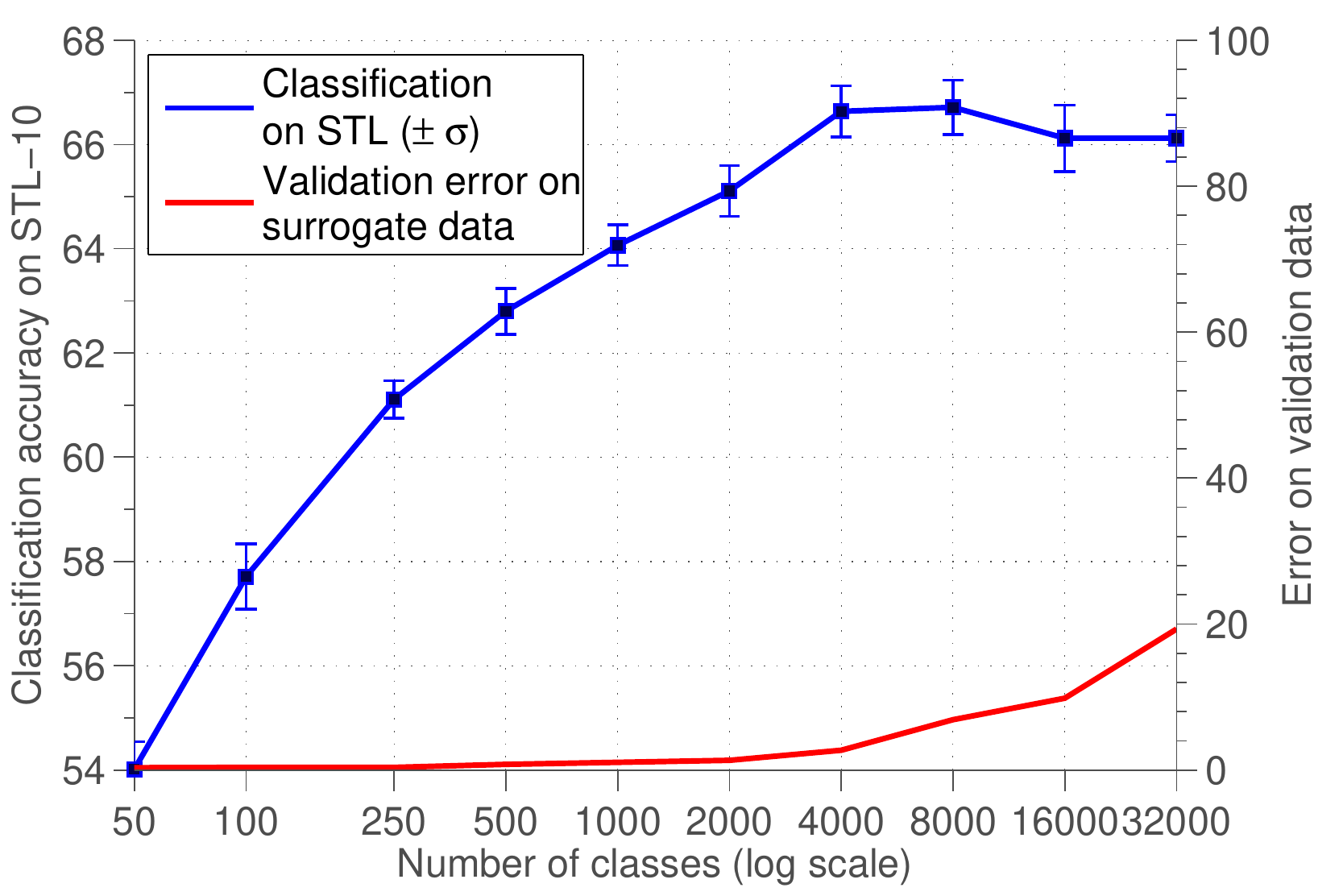}
  \begin{minipage}[t]{.9\linewidth}
  \caption{\label{fig:var_num_classes}Dependence of classification
    accuracy on STL-10 on the number of surrogate classes in the
    training data. For reference, the error on validation surrogate
    data is also shown. Note the different scales for the two graphs.}
  \end{minipage}
\end{minipage}%
\begin{minipage}[t]{.5\textwidth}
  \centering
  \includegraphics[width=.95\linewidth]{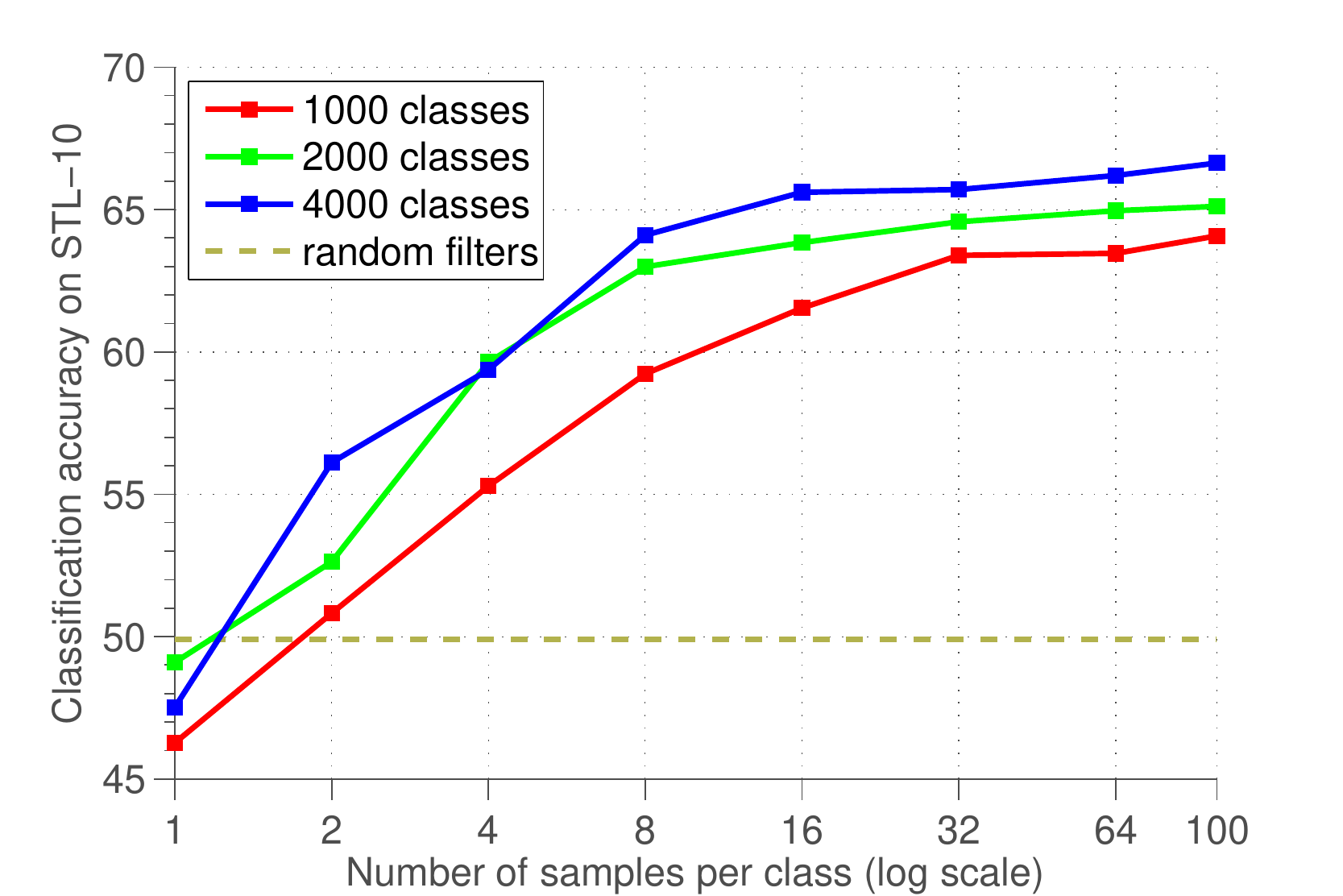}
  \begin{minipage}[t]{.9\linewidth}
  \caption{\label{fig:var_num_samples}Dependence of classification
    accuracy on STL-10 on the number of samples per surrogate
    class. Standard deviations not shown to avoid clutter.}
  \end{minipage}
\end{minipage}
\end{figure}

Our pipeline lets us easily vary the number of surrogate classes in the training data and the number of training samples per surrogate class. We use this to measure the effect of these factors on the quality of the resulting features. We vary the number of surrogate classes between $50$ and $32000$ and the number of training samples per surrogate class between $1$ and $100$. The results are shown in Fig.~\ref{fig:var_num_classes} and ~\ref{fig:var_num_samples}. In Fig.~\ref{fig:var_num_samples} we also show, as a baseline, the classification performance of random filters (all weights are sampled from a normal distribution with standard deviation $0.001$, all biases are set to zero). Initializing the random filters does not require any training data and can hence be seen as using $0$ samples per surrogate class. Error bars in Fig.~\ref{fig:var_num_classes} show the standard deviations computed when testing on $10$ folds of the STL-10 dataset. 
 
An apparent trend in Fig.~\ref{fig:var_num_classes}  is that increasing the number of surrogate classes results in an increase in classification accuracy until it reaches an optimum at around $8000$ surrogate classes. When the number of surrogate classes is further increased the classification results do not change or slightly decrease. One explanation for this behavior is that the larger the number of surrogate classes becomes, the more these classes overlap. As a result of this overlap the classification problem becomes more difficult and adapting the network to the surrogate task no longer succeeds. To check the validity of this explanation we also plot in Fig.~\ref{fig:var_num_classes} the classification error on the validation set (taken from the surrogate data) computed after training the network.  It rapidly grows as the number of surrogate classes increases, supporting the claim that the task quickly becomes more difficult as the number of surrogate classes increases. 

Fig.~\ref{fig:var_num_samples} shows that classification accuracy
increases with increasing number of samples per surrogate class and
saturates around $100$ samples. It can also be seen that when training
with small numbers of samples per surrogate class, there is no clear
indication that having more classes lead to better performance. We
hypothesize that the reason may be that with few training samples per
class the surrogate classification problem is too simple and hence the
network can severely overfit, which results in poor and unstable generalization to real classification tasks. However, starting from around $8-16$ samples per surrogate class, the surrogate task gets sufficiently complicated and the networks with more diverse training data (more surrogate classes) perform consistently better.

\section{Discussion}
We proposed a simple unsupervised feature learning approach based on data augmentation that shows good results on a variety of classification tasks. While our approach sets the state of the art on STL-10 it remains to be seen whether this success can be translated into consistently better performance on other datasets.


The performance of our method saturates when the number of surrogate classes increases. One probable reason for this is that the surrogate task we use is relatively simple and does not allow the network to learn complex invariances such as 3D viewpoint invariance or inter-instance invariance. We hypothesize that our unsupervised feature learning method could learn more powerful higher-level features if the surrogate data were more similar to real-world labeled datasets. This could be achieved by using extra weak supervision provided for example by video data or a small number of labeled samples. Another possible way of obtaining richer surrogate training data would be (unsupervised) merging of similar surrogate classes. We see these as interesting directions for future work.

\section*{Acknowledgements}
We acknowledge funding by the ERC Starting Grant VideoLearn (279401).

{\small
\bibliographystyle{ieee}
\bibliography{../dosovits.bib}
}

\end{document}